\documentclass{article}
\usepackage{amsmath,epsfig}

\title{Variational local structure estimation for image super-resolution}
\author{Heng Lian\\Division of Applied Mathematics, Brown University\\Providence, RI 02912}
\begin{document}
%\ninept
%
\maketitle
\begin{abstract}
Super-resolution is an important but difficult problem in image/video processing. If a video sequence or some training set other than the given low-resolution image  is available, this kind of extra information can greatly aid in the reconstruction of the high-resolution image. The problem is substantially more difficult with only a single low-resolution image on hand. The image reconstruction methods designed primarily for denoising is insufficient for super-resolution problem in the sense that it tends to oversmooth images with essentially no noise. We propose a new adaptive linear interpolation method based on variational method and inspired by local linear embedding (LLE). The experimental result shows that our method avoids the problem of oversmoothing and preserves image structures well. 
\end{abstract}


\section{Introduction}
\label{sec:intro}

Super-resolution is an image processing technique to obtain high-resolution image from its low-resolution counterpart, either using information from a single image or  image sequence. Most of the literature focus on the latter \cite{he}\cite{schultz}. With only one image, it is generally difficult to interpolate the missing pixels with little prior information about the image. 

Many approaches have been proposed in recent years. For video/image sequence, information on subpixel motion between temporally adjacent images can be utilized. Due to the ill-posedness of the super-resolution problem, regularization\cite{he}  or a Bayesian framework\cite{woods} usually needs to be set up, the two approaches being closely related.  

For only one single image, traditionally the most often used method is bilinear or bicubic interpolation, which works under the assumption that images are smooth for the most part. One recent advance on single image super-resolution is the work by Freeman et al\cite{freeman}. Using a large database of image patches consisting of low-resolution and high-resolution pairs, the algorithm will search the database for the corresponding high-resolution patch, for each  of the image patches in the given low-resolution image. One obvious disadvantage of this approach is that it requires the existence of a database in the first place.

We propose a novel algorithm for single image super-resolution, without the need for a database or other training images. Our algorithm is inspired by Local Linear Embedding (LLE) used for dimension reduction. By trying to express each pixel value as a linear combination of neighboring pixel values, we can estimate the local structure of the low-resolution image which we can then use to construct a filter for interpolation in the second step. 

One work similar in spirit to ours is \cite{adrian}. There the authors also try to express the center pixel as a linear combination of its neighbors. Similar to Freeman\cite{freeman}, their algorithm also requires a training set of pairs of patches in order to estimate those linear coefficients. And the coefficients are learned in the regression framework, giving as a result a spatially adaptive filter. There algorithm can be regarded as a complement to \cite{freeman} which works well on cartoon images.

One approach to single image super-resolution without training, which is also related to ours, is mentioned in \cite{chanshen} using digital TV filter\cite{chenoshershen}, which is the discrete counterpart of the variational  functional method with a total variation (TV) norm. The digital TV filter is originally designed with the goal of image denoising in mind. Although the authors demonstrated that using a TV smoothing term can largely preserve edges and other important features, the result still looks blurred and blocky (using higher order smoothness term can alleviate the second problem with the cost of exacerbating the first problem). We think that directly using denoising algorithm on the super-resolution problem is somewhat unnatural, with no special attention to the problem at hand. Our algorithm uses variational approach on the linear coefficients instead, so the smoothing term has an effect on the local filter, not directly on the image.

\section{local structure estimation}
\label{sec:estimation}
Denoting the given image by $u$, defined  on $\Omega=[0,1]\times[0,1]$.
We start by assuming that each point $x\in\Omega$ can be approximately expressed as an weighted combination of its neighbors (eight neighbors are used in this paper):
\begin{equation}\label{e1}
u(x)\approx \sum_{i=1}^8 w^{(i)}(x)u(x+\delta_i)
\end{equation}
, where $x+\delta_i$ are the neighbors of $x$, with $\delta_1=(-h,-h)$ etc. For grayscale images, for example, the above equality is trivially satisfied for an infinite number of combinations of $w^{(i)}, i=1..8$, given the image $u$. So trying to estimate $w$ separately for each point $x$ is futile without further constraint. Here we use the variational approach with a TV norm proposed in \cite{rudin}:
\begin{equation}\label{e2}
\min_w\int\left(\sum_i w^{(i)}u(\cdot+\delta_i)-u\right)^2dx
+\lambda\sum_{i=1}^8\int|\nabla w^{(i)}|dx
\end{equation}
,where $\lambda$ is a parameter that must be set to balance the fidelity term and the smoothness term. 
This minimization problem can be solved iteratively using the corresponding evolution PDE:
\begin{equation}\label{e3}
w_t^{(i)}=div\left(\frac{\nabla w^{(i)}}{|\nabla w^{(i)}|}\right)-\lambda\left(\sum_iw^{(i)}u(\cdot+\delta_i)-u\right)u(\cdot+\delta_i)\,
\end{equation}
,which can then be discretized and solved numerically.

\begin{figure}[!htb]
\begin{minipage}[b]{1.0\linewidth}
  \centering
  \centerline{\epsfig{figure=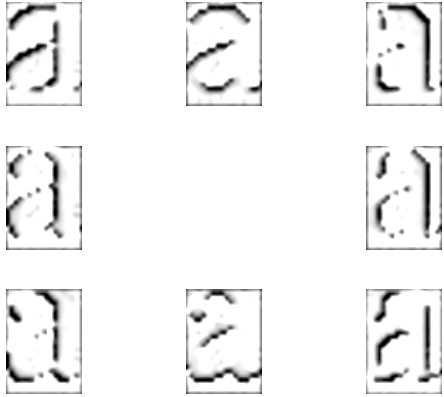,width=8.5cm}}
  \centerline{(a) $w^{(i)}$ computed in our method}\medskip
\end{minipage}
\begin{minipage}[b]{1.0\linewidth}
  \centering
  \centerline{\epsfig{figure=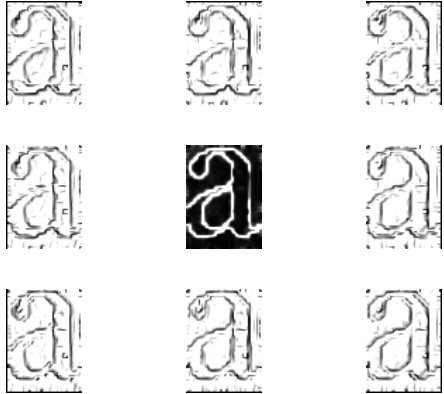,width=8.5cm}}
  \centerline{(b) $h^{(i)}$ for TV filter after convergence}\medskip
\end{minipage}
\caption{The linear coefficients computed in our method and digital TV filter is quite different for a simple image}
\label{fig:res}
\end{figure}

\subsection{Connections to LLE}
\label{ssec:subhead}
Given a set of $N$ vectors $\{X_i\}$ in high dimensions. The LLE solves the following minimization problem:
\begin{equation}\label{e5}
\min_W \sum_i (X_i-\sum_jW_{ij}X_j)^2
\end{equation}
with the constraint
\begin{equation}\label{e6}
 \sum_j W_{ij}=1,W_{ij}\ge 0 
\end{equation}
where the index $j$ runs over those of $X_j$ that is among the $k$ nearest neighbors of $X_i$. 
with nearest neighbors determined by some metric $d(X_i,X_j)$. This minimization problem has a non-trivial solution since it is usually assumed that the dimension of $X_i$ is much bigger than $k$.

To generalize LLE, we first assume that the data come in with two components $X_i=(Y_i,Z_i)$ (think of $Y_i$ as grid position, and $Z_i$ as intensity value). Now we can minimize the following:
\begin{equation}\label{e7}
\min_W \sum_i (Z_i-\sum_jW_{ij}Z_j)^2
\end{equation}
 the index $j$ still runs over $k$ nearest neighbors of $X_i$. 
but now with nearest neighbors determined by some metric $d(Y_i,Y_j) $ 
depending on the other component of $X$.

If dimension of $X_i$ is small compared to $k$ (as in the case of an image), we must add regularization term to make the problem well-posed. And we will recover the discrete counterpart of (\ref{e2}) after ignoring the convexity constraint(\ref{e6}).

\subsection{Comparison with Digital TV Filter}
\label{ssec:tv}
The digital TV filter also implicitly estimates some kind of local linear structure. It repeatedly applies the following filtering step on the current image until convergence ( see equation $(9)$ in \cite{chenoshershen}):
\begin{equation}\label{e4}
u^{t+1}(x)=\sum_{i=1}^8 h^{(i)}(x)u^t(x+\delta_i)+h^{(0)}(x)u^0(x)
\end{equation}

, where $u^t$ is the reconstructed image at  the $t$-th iteration, and the coefficients $h$ are based on local image gradient and must be recomputed for each iteration. Note one difference in form between (\ref{e1}) and (\ref{e4}) is that we don't have a corresponding $w^{(0)}$ in (\ref{e1}). The reason is that if $w^{(0)}$ were added, (\ref{e2}) would have a trivial solution, with $w^{(0)}\equiv 1$, $w^{(i)}\equiv 0, i=1\ldots 8$. The coefficients for digital TV filter after convergence is quite different from the coefficients estimated from our method (Fig. 1). This is due to the fact that we estimate the coefficients using a global functional, and the TV filter computes the coefficients using local information only. Another reason for the difference is that our approach estimates orientation-specific weight, the digital TV only estimates local gradient magnitude.

\section{super-resolution}
\label{sec:pagestyle}
Suppose now we are given an $n\times n$ low-resolution image $u^{low}$, and have also obtained an initial reconstructed $zn\times zn$ high-resolution image $\hat{u}_0$ (by nearest neighbor interpolation, for example), where $z$ is the magnification ratio. Using variational method outlined in the last section on $u^{low}$, we get eight $n\times n$ weight matrices $w^{(i)}$ representing roughly the local structure in the low-resolution image. These weight matrices cannot be directly used as filters for high-resolution image because it is smaller in size. To obtain suitable $zn\times zn$ weight matrices $\tilde{w}^{(i)}$, we can use linear interpolation on the low-resolution weight matrices to obtain the high-resolution weight matrices. Finally, the weight matrices can act as adaptive filters for interpolation:
\begin{equation}\label{e8}
u^{high}(x)= \sum_{i=1}^8 \tilde{w}^{(i)}(x)\hat{u}_0(x+\delta_i)
\end{equation}

\section{experimental result}
\label{sec:typestyle}

We compare the proposed method with both bicubic interpolation and digital TV filter using grayscale images.  We use $z=3$ in the following experiments. The smoothing parameter used is $\lambda=0.1$ in all our experiments. It is observed that the result is quite similar for all different smoothing parameters we tried. The reason is probably that the smoothness parameter affect the smoothness of the weights, and only has indirect effect on image itself through the weights. We first apply the algorithm to the house image (Fig. 2). Both bicubic interpolation and digital TV filter oversmooth the image. To make this point clearer, the Sobel edge detector is applied on the results using the same threshold. The edge detector returns less edges on both bicubic interpolated and TV filtered images, which shows the oversmoothing problem of both methods. Another advantage of our method over bicubic interpolation is that the edges are less wiggled, as can be seen from the roof of the house image in (c) of Fig 2, for example.

 As a second example, the fingerprint image is used to see how well our method works for texture images. It is well known that TV filter does not work on texture images as can be seen from Fig 3. Our method also gives reasonable result in this case.

% Below is an example of how to insert images. Delete the ``\vspace'' line,
% uncomment the preceding line ``\centerline...'' and replace ``imageX.ps''
% with a suitable PostScript file name.
% -------------------------------------------------------------------------

%%%%%%%%%%%%%%%%%%%%%%%%%%%%%%%
%%%%%%%% my own figures %%%%%%%
%%%%%%%%%%%%%%%%%%%%%%%%%%%%%%%
%\begin{figure}[htb]
%\begin{minipage}[b]{1.0\linewidth}
%  \centering
%  \centerline{\epsfig{figure=Whouse.eps,width=8.5cm}}
%  \centerline{(a) Result 1}\medskip
%\end{minipage}
%\begin{minipage}[b]{1\linewidth}
%  \centering
%  \centerline{\epsfig{figure=Hhouse.eps,width=8.5cm}}
%  \centerline{(b) Results 3}\medskip
%\end{minipage}
%\caption{Example of placing a figure with experimental results.}
%\label{fig:res}
%\end{figure}

\begin{figure}[htb]
\begin{minipage}[b]{1.0\linewidth}
  \centering
  \centerline{\epsfig{figure=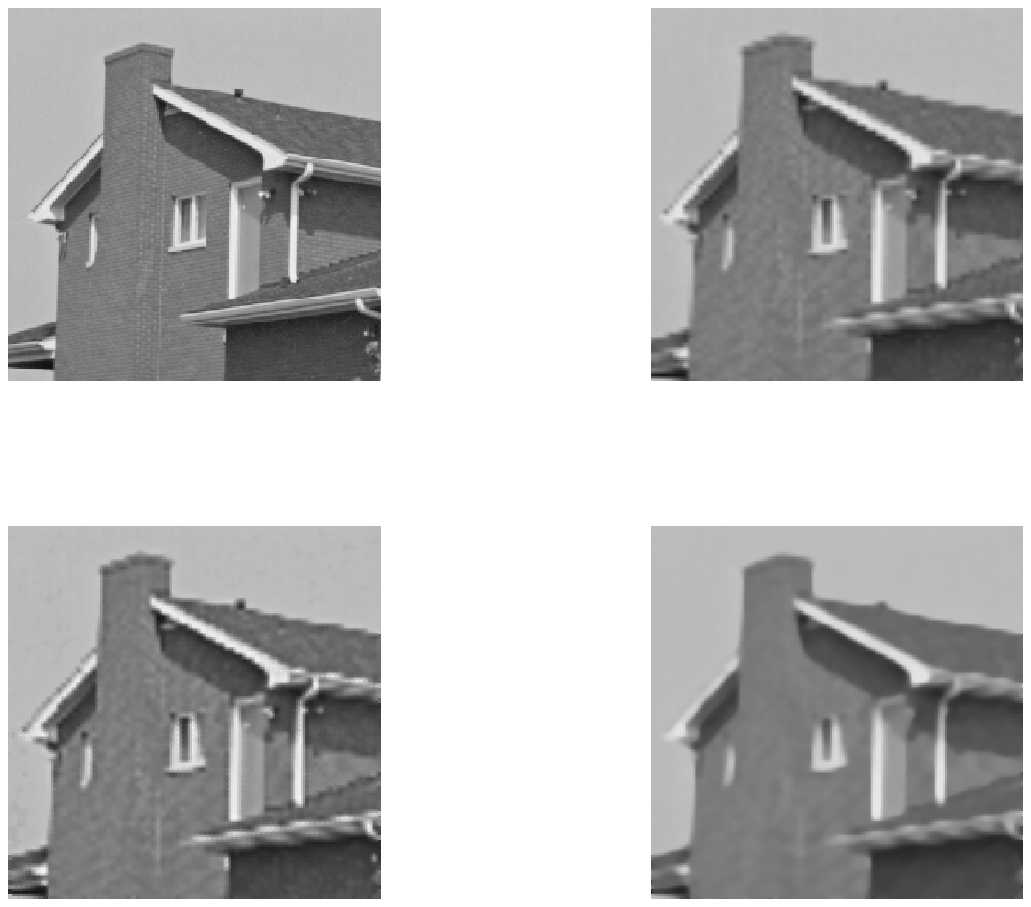,width=8.5cm}}
  \centerline{(a) interpolation results }%\smallskip
\end{minipage}
\begin{minipage}[b]{1.0\linewidth}
  \centering
  \centerline{\epsfig{figure=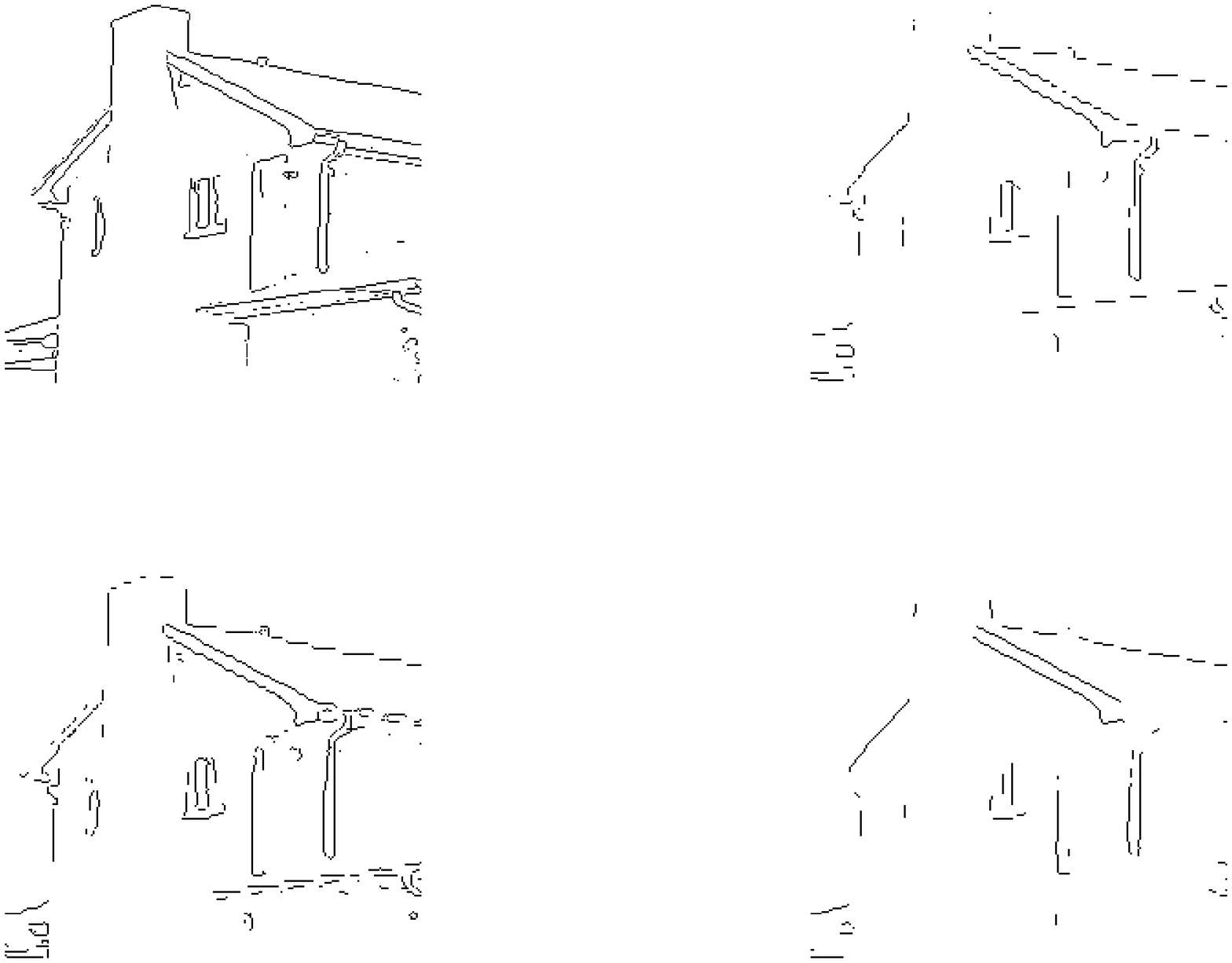,width=8.5cm}}
  \centerline{(b) response to Sobel edge detector}%\smallskip
\end{minipage}
\begin{minipage}[b]{1.0\linewidth}
  \centering
  \centerline{\epsfig{figure=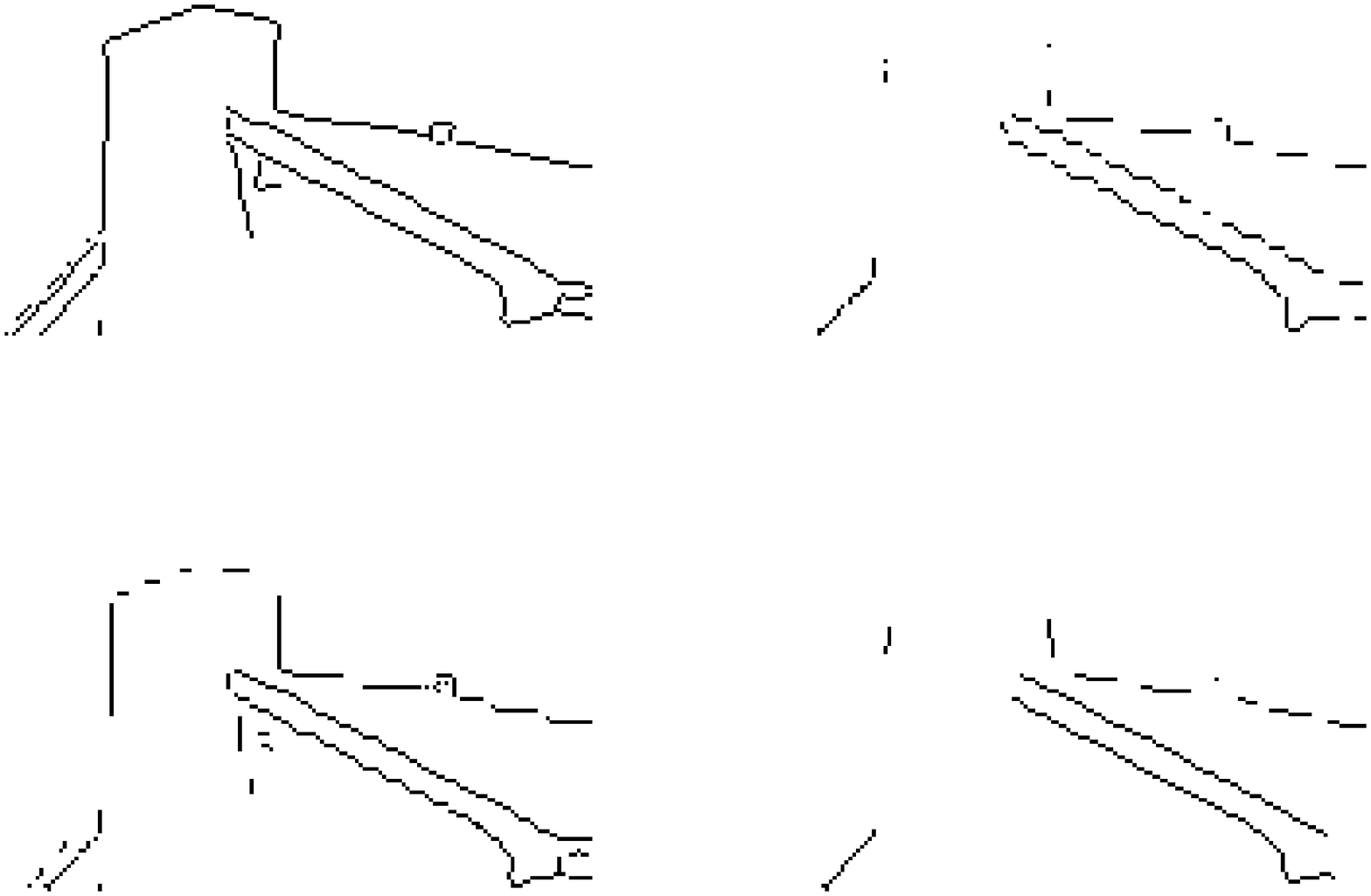,width=8.5cm}}
  \centerline{(c) zoom-in of part in (b) }%\smallskip
\end{minipage}
\caption{Comparison of interpolation results. Upper Left: original high-resolution image. Upper Right: bicubic interpolation, PSNR=26.9. Lower Left: our method, PSNR=27.6. Lower Right: digital TV filter, PSNR=26.9.}
\label{fig:res}
\end{figure}

\begin{figure}[htb]
\begin{minipage}[b]{1.0\linewidth}
  \centering
  \centerline{\epsfig{figure=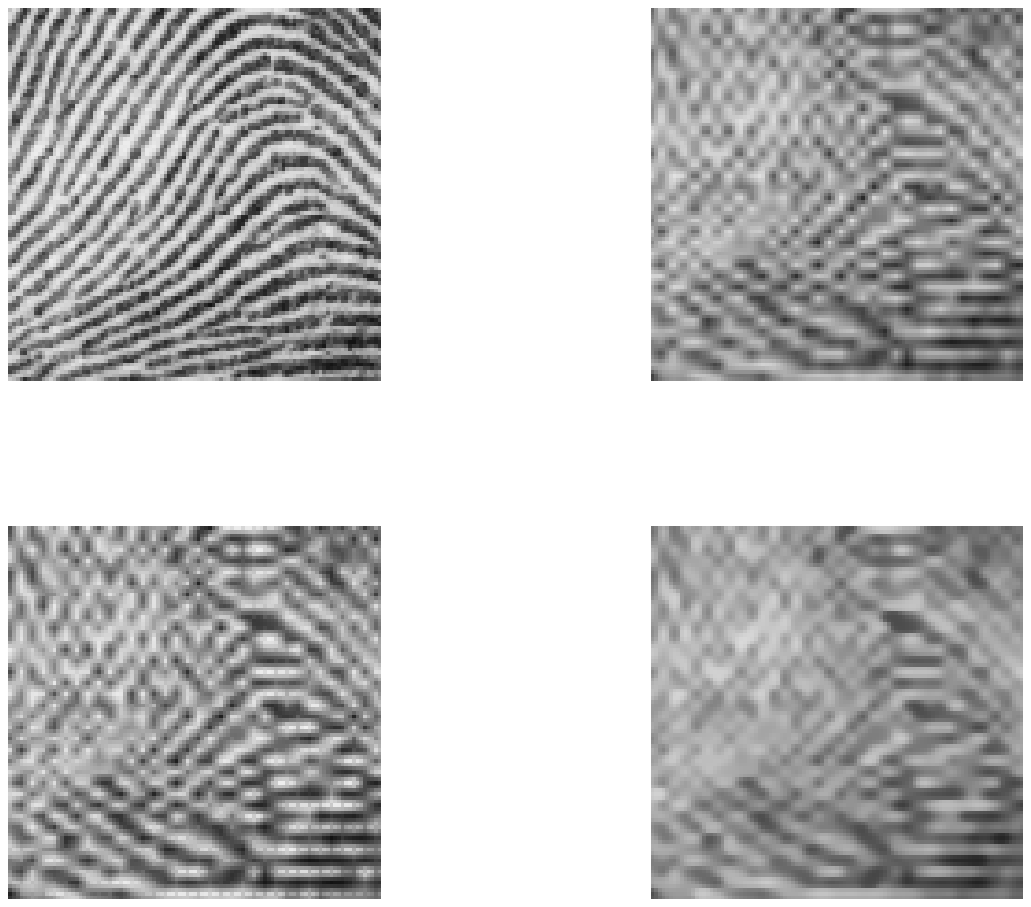,width=8.5cm}}
  \centerline{(a) interpolation results}\medskip
\end{minipage}
\begin{minipage}[b]{1.0\linewidth}
  \centering
  \centerline{\epsfig{figure=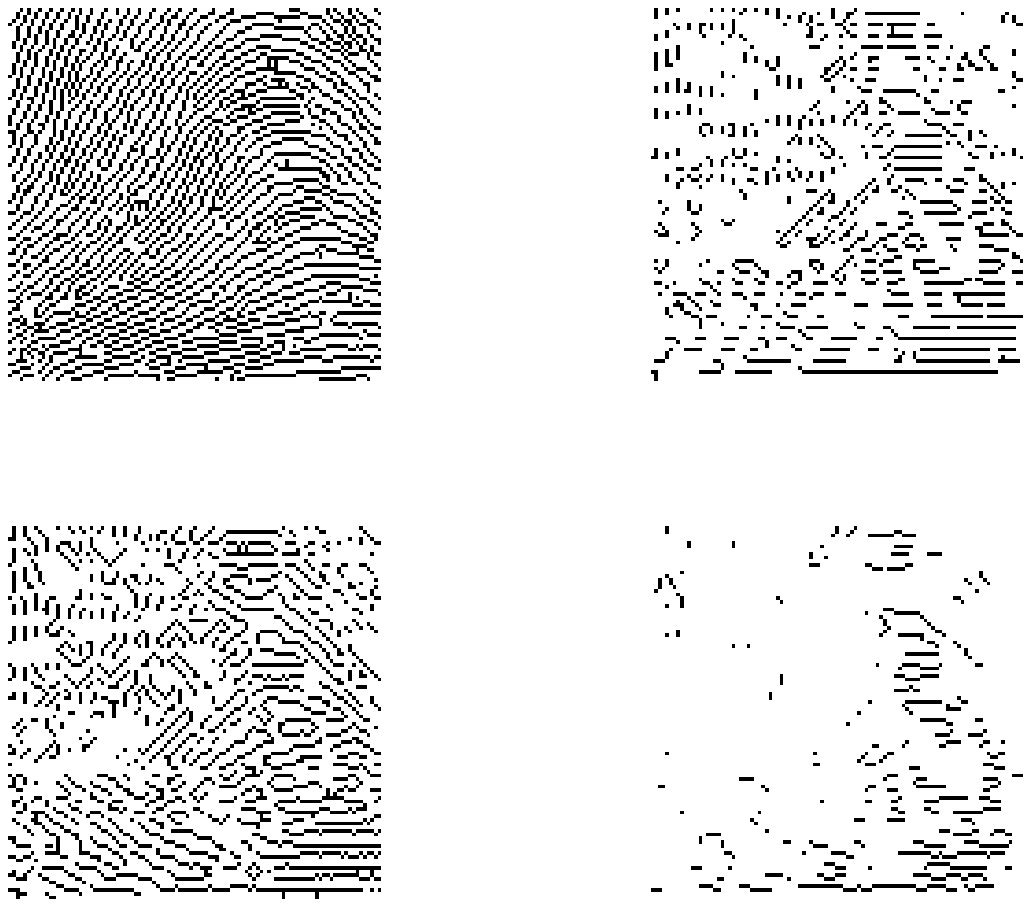,width=8.5cm}}
  \centerline{(b) response to Sobel edge detector}\medskip
\end{minipage}
\caption{Comparison of results on fingerprint image. Upper Left: original high-resolution image. Upper Right: bicubic interpolation, PSNR=16.3. Lower Left: our method, PSNR=16.8. Lower Right: digital TV filter, PSNR=16.2.}
\label{fig:res}
\end{figure}

% To start a new column (but not a new page) and help balance the last-page
% column length use \vfill\pagebreak.
% -------------------------------------------------------------------------
%\vfill
%\pagebreak

\section{Conclusion}
\label{sec:conclusion}
In this work, we proposed a new algorithm for single-image super-resolution problem using variational method. Instead of working on the image space as in the previous work utilizing variational method, we use variational formulation to estimate the local structure of an image. The resulting adaptive filter reflects both local pixel variance and global image information. The experimental result shows some advantage of our method over some previous approaches. A future research direction might be to explore other applications of the variational estimation of the local image structure.

% References should be produced using the bibtex program from suitable
% BiBTeX files (here: strings, refs, manuals). The IEEEbib.bst bibliography
% style file from IEEE produces unsorted bibliography list.
% -------------------------------------------------------------------------
\bibliographystyle{IEEEbib}
\bibliography{strings,refs}

\end{document}